\documentclass[conference]{IEEEtran}
\IEEEoverridecommandlockouts
\usepackage{cite}
\usepackage{amsmath,amssymb,amsfonts}
\usepackage{algorithmic}
\usepackage{graphicx}
\usepackage{textcomp}
\usepackage{xcolor}
\usepackage{multirow}
\usepackage{hyperref}
\def\BibTeX{{\rm B\kern-.05em{\sc i\kern-.025em b}\kern-.08em
    T\kern-.1667em\lower.7ex\hbox{E}\kern-.125emX}}
\begin{document}

\title{Data Efficiency and Transfer Robustness in Biomedical Image Segmentation: A Study of Redundancy and Forgetting with Cellpose}

\author{\IEEEauthorblockN{1\textsuperscript{st} Shuo Zhao}
\IEEEauthorblockA{
\textit{Leibniz Institut für Analytische Wissenschaften – ISAS – e.V.}\\
Dortmund, Germany \\
shuo.zhao@isas.de}
\and
\IEEEauthorblockN{2\textsuperscript{nd} Jianxu Chen}
\IEEEauthorblockA{
\textit{Leibniz Institut für Analytische Wissenschaften – ISAS – e.V.}\\
Dortmund, Germany \\
jianxu.chen@isas.de}
}

\maketitle

\begin{abstract}
Generalist biomedical image segmentation models such as Cellpose are increasingly applied across diverse imaging modalities and cell types. However, two critical challenges remain underexplored: (1) the extent of training data redundancy and (2) the impact of cross domain transfer on model retention. In this study, we conduct a systematic empirical analysis of these challenges using Cellpose as a case study. Firstly, to assess data redundancy, we propose a simple dataset quantization (DQ) strategy for constructing compact yet diverse training subsets. Experiments on the Cyto dataset show that image segmentation performance saturates with only 10\% of the data, revealing substantial redundancy and potential for training with minimal annotations. Latent space analysis using MAE embeddings and t-SNE confirms that DQ selected patches capture greater feature diversity than random sampling. Secondly, to examine catastrophic forgetting, we perform cross domain finetuning experiments and observe significant degradation in source domain performance, particularly when adapting from generalist to specialist domains. We demonstrate that selective DQ based replay reintroducing just 5–10\% of the source data effectively restores source performance, while full replay can hinder target adaptation. Additionally, we find that training domain sequencing improves generalization and reduces forgetting in multi-stage transfer. Our findings highlight the importance of data-centric design in biomedical image segmentation and suggest that efficient training requires not only compact subsets but also retention aware learning strategies and informed domain ordering. The code is available at \href{https://github.com/MMV-Lab/biomedseg-efficiency}{https://github.com/MMV-Lab/biomedseg-efficiency}.

\end{abstract}


\section{Introduction}

Instance image segmentation is an important task in biomedical image analysis. Recently, Cellpose~\cite{stringer2021cellpose} stands out as a generalist framework that delivers strong performance across diverse cell types and imaging modalities. However, there are two underexplored yet critical concerns: (1) how much of its large training set is truly necessary, and (2) how robustly it retains knowledge when adapted to new domains~\cite{cao2025rethinking}.

Why should we care about redundancy? Because pixel level annotations are expensive and time consuming. If only 10\% of the data is sufficient to train an effective model, biomedical teams could potentially save 90\% of the labeling workload and computational resources.  

Why should we care about forgetting? Image segmentation models are increasingly reused across laboratories, datasets, and imaging modalities. A model that forgets previously acquired knowledge after each finetuning round becomes unreliable, unscalable, and may ultimately no longer qualify as a general model. These challenges remain poorly quantified in the biomedical image analysis literature.

In this paper, we empirically investigate both issues using the Cellpose framework as a testbed. We are \textbf{not} critiquing Cellpose. On the contrary, it is precisely because Cellpose is so powerful, general, and widely deployed that we have chosen it as a case study. Our goal is diagnostic rather than architectural: we do not propose a new model, but rather expose the hidden inefficiencies and fragilities of existing training and adaptation workflows. We employ dataset quantization (DQ) to evaluate redundancy, and use cross domain finetuning to examine forgetting. We further extend the study to the NeurIPS 2022 Cell Segmentation Challenge dataset~\cite{NeurIPS-CellSeg}, which features diverse modalities and institutions, enabling a realistic stress test of cross domain generalization.

We also explore multi-stage transfer paths to analyze how the sequence of domain exposure affects performance. We find that simpler domains are easier to forget, while complex domains benefit from being placed earlier or intermediately in the training sequence. Our findings provide both practical insights and theoretical implications for building scalable, memory aware biomedical image segmentation systems.

\subsection*{Our contributions are:}
\begin{itemize}
    \item We show that about 10\% of the Cellpose training data yields near saturated performance with high redundancy.
    \item Finetuning Cellpose on MoNuSeg leads to catastrophic forgetting, with source performance collapsing.
    \item We validate both phenomena under more challenging cross domain conditions using the NeurIPS 2022 dataset.
    \item We analyze zero-shot and sequential transfer settings, highlighting the importance of dataset composition and adaptation order.
    \item We provide a holistic analysis of Cellpose’s training behavior from a data-centric, transfer aware perspective.
\end{itemize}

\section{Related Work}

This section reviews prior research on generalist biomedical image segmentation models, data efficient training, and forgetting in transfer learning.

\subsection{Generalist Biomedical Image Segmentation Models}

Cellpose and its successors~\cite{stringer2021cellpose, pachitariu2022cellpose, stringer2025cellpose3, pachitariu2025cellpose} represent a major shift in biomedical instance image segmentation with a flow based approach. Cellpose 1.0 introduced a robust generalist trained on diverse datasets. Cellpose 2.0 enabled user specific retraining with minimal data, and Cellpose 3.0 integrated denoising and SAM prompting for improved robustness. These tools now form the backbone of image segmentation workflows in many biomedical laboratories and software.

Despite their generality, the data demands and transfer behavior of Cellpose models remain poorly understood. Official pipelines assume full data retraining, and the contribution of training samples remains largely unquantified. Moreover, finetuning across datasets, commonly done in practice, is rarely evaluated for its retention properties. We aim to fill this gap with a data-centric evaluation of training efficiency and knowledge transfer.

\subsection{Dataset Quantization and Coreset Selection}

Dataset quantization (DQ) is a coreset selection strategy that partitions the feature space and selects representative samples to reduce dataset size while preserving performance. Traditional coreset selection methods emphasized geometric or visual diversity, while recent work utilizes representativeness~\cite{zhao2025active}, entropy~\cite{nath2020diminishing}, uncertainty~\cite{gaillochet2023active}, or gradient based embeddings~\cite{wang2024comprehensive} for more principled selection. Biomedical image segmentation poses unique challenges: sparse ground truth, high annotation cost, and multi-modal distributions.

Although recent studies show promise~\cite{yang2023towards, hong2024evolution}, most approaches have not been tested in pretrained models like Cellpose. Our work revisits the DQ as a simple, interpretable baseline to investigate redundancy, performance saturation, and feature space coverage.

\subsection{Catastrophic Forgetting in Biomedical Transfer}
Catastrophic forgetting describes how a model “forgets” previously learned knowledge when adapting to new data. Continual learning research has long identified catastrophic forgetting as a key limitation in sequential task learning~\cite{qazi2024continual}.  Biomedical models are often adapted across domains, such as from fluorescence to histopathology, and forgetting can undermine reliability. However, the dominant practice remains naive finetuning often without retention aware evaluation~\cite{yoon2023domain, garg2022multi}.

Recent biomedical studies highlight the asymmetry in forgetting~\cite{qu2025recent}: models tend to forget general tasks more severely when adapted to narrow domains. Our experiments confirm this behavior and further quantify the role of training order, DQ subset composition, and zero-shot generalization. These insights pave the way for future work on memory augmented, modular, or rehearsal architectures in real-world biomedical image segmentation pipelines.

\section{Methodology}\label{sec:methodology}

Our methodology is structured to investigate two central questions in biomedical image segmentation empirically: 
(1) How redundant the Cellpose training data is and
(2) How severely finetuning causes forgetting across domains.
We approached the problem empirically, exploring the effectiveness of simple replay mechanisms and the impact of different domain training orders on forgetting behavior.

\subsection{Datasets}

We conducted experiments on three public datasets:

\textbf{Original Cellpose Dataset}: Comprising 540 annotated microscopy images which include a wide range of cell types and imaging modalities. It serves as our primary domain for training, evaluation, and subset quantization.

\textbf{MoNuSeg Dataset}~\cite{kumar2019multi}: A histopathology benchmark consisting of 37 H\&E stained tissue images with pixel precise nuclear annotations. MoNuSeg dataset is used as a representative domain specific target in transfer experiments.

\textbf{NeurIPS 2022 Cell Segmentation Dataset}~\cite{NeurIPS-CellSeg}: A large-scale, multi-site benchmark with diverse modalities, such as fluorescence and CODEX, offering high intra-class variability for testing zero-shot generalization and sequential transfer.

For clarity, we denote the Cellpose dataset as \textbf{Cyto}, MoNuSeg as \textbf{Histo}, and NeurIPS 2022 as \textbf{MultiInst}, based on their modality and content to avoid naming confusion.

We only process labeled data. We evaluated instance image segmentation performance using IoU, dice, precision, recall, accuracy, and panoptic quality on both source and target testsets to assess generalization, forgetting, and adaptation.

\subsection{Dataset Quantization for Coreset Selection}

To identify compact yet informative training subsets, we adopt Dataset Quantization (DQ)~\cite{zhou2023dataset}, a bin based coreset selection framework. The process involves:

\textbf{Feature Extraction:} We extract patch features using a pretrained Masked Autoencoder (MAE)~\cite{he2022masked}. These features are used to represent patch diversity in a high dimensional latent space.

\textbf{Bin Formation:} Using pairwise feature distances, the dataset is partitioned into $N$ non overlapping bins $\{S_1, ..., S_N\}$. Within each bin, samples are selected based on a submodular gain criterion:

\begin{equation}
P(x_k) = \sum_{p \in S^{k-1}_n} \|f(p) - f(x_k)\|^2 - \sum_{p \in D \setminus S^{k-1}_n} \|f(p) - f(x_k)\|^2,
\label{eq:submodular_gain}
\end{equation}

where $f(\cdot)$ denotes the MAE features, $S^{k-1}_n$ is the partial bin, and $D$ is the full patch set.

\textbf{Subset Sampling:} A fixed proportion $\rho$ is uniformly sampled from each bin to construct the final coreset:

\begin{equation}
S^* = \bigcup_{n=1}^{N} g(S_n, \rho),
\end{equation}

ensuring that feature diversity is preserved even under high dataset compression.

\subsection{Training and Preprocessing Details}

We used the official Cellpose training interface. All the experiments were run on a single NVIDIA A100 GPU (40GB). 

Patch Extraction: All images are patched using a $224 \times 224$ sliding window with stride 112, resulting in 10,063 (Cellpose Cyto), 2,997 (MoNuSeg Histo), and 128,458 (NeurIPS MultiInst) labeled patches.

Training Hyperparameters: grayscale input ($\text{chan} = 0$), learning rate = 0.1, weight decay = $1 \times 10^{-4}$, epochs = 500, checkpoint every 50 epochs.

\subsection{Zero-Shot Generalization}

To assess cross domain robustness, we perform zero-shot inference on the MultiInst testset using Cellpose models trained exclusively on the Cyto dataset, either the full 100\% or a compressed 30\% DQ subset. This setup evaluates how well models generalize to unseen and heterogeneous modalities without any target domain exposure and whether training on compact, curated subsets compromises transferability.

\subsection{Cross Domain Finetuning and Forgetting Analysis}

To assess how models retain knowledge across domains, we conducted two key finetuning experiments.

\begin{itemize}
    \item \textbf{Cyto $\rightarrow$ Histo}: A model trained on Cellpose trainset is finetuned on MoNuSeg trainset, and its performance on Cyto is double evaluated to quantify forgetting.
    \item \textbf{Histo $\rightarrow$ Cyto}: The reverse setup, testing the asymmetry of knowledge loss.
\end{itemize}

No memory preserving mechanisms such as rehearsal or regularization are used, allowing us to isolate the effects of naive finetuning.

To explore lightweight mitigation of forgetting, we also design replay experiments by combining a small proportion of original training data (e.g., 1–50\% of Cyto) with full Histo finetuning. This simulates joint training under resource constraints and tests whether selective retention improves performance stability.

\subsection{Multi-Stage Sequential Transfer}

To investigate how the order of sequential domain adaptation affects both generalization and forgetting, we design controlled experiments across three representative multi-stage transfer paths using the \textbf{Cyto}, \textbf{Histo}, and \textbf{MultiInst} datasets.

\begin{itemize}
    \item \textbf{Path A:} Cyto $\rightarrow$ Histo $\rightarrow$ MultiInst
    \item \textbf{Path B:} Cyto $\rightarrow$ MultiInst $\rightarrow$ Histo
    \item \textbf{Path C:} MultiInst $\rightarrow$ Cyto $\rightarrow$ Histo
\end{itemize}

These three paths are selected from the six possible permutations to strike a balance between experimental feasibility and conceptual coverage. Specifically, they capture three key hypotheses: (1) whether generalist domains like Cyto are more vulnerable to forgetting when trained first (Path A, B); (2) whether diverse datasets like MultiInst provide stronger initializations when placed at the beginning (Path C); and (3) how final-domain performance varies when specialized datasets like Histo are placed last (Path B, C). The remaining permutations are excluded as they offer redundant or less informative variations of these scenarios.

Each model is evaluated on all three domains to assess:
\begin{itemize}
    \item How adaptation order affects forgetting and knowledge recovery.
    \item Whether starting from a diverse base (e.g., MultiInst) improves retention and transfer.
    \item Whether DQ-trained models provide more stable multi-domain performance.
\end{itemize}

This experiment simulates real-world biomedical workflows, where models are incrementally adapted across heterogeneous datasets and institutions in non-linear order. Understanding the impact of training sequence is thus critical for building robust, reusable image segmentation pipelines.

\subsection{Simple Replay Strategy for Forgetting Mitigation}
To mitigate catastrophic forgetting during cross domain finetuning, we implement a simple replay strategy that reintroduces a small portion of the source domain data during training on the target domain.
This approach allows the model to retain critical source knowledge while adapting to the new domain, without requiring complex multitask or continual learning frameworks.
While this is not the most sophisticated solution to catastrophic forgetting, we present it as a lightweight and interpretable baseline for empirical analysis.
This strategy is motivated by experience replay techniques~\cite{rolnick2019experience}.

\section{Results and Analysis}

\begin{table*}[t]
\caption{Image segmentation Performance at Different Quantization Rates in Cyto Dataset (Mean ± Std)}
\begin{center}
\begin{tabular}{|c|c|c|c|c|c|c|}
\hline
\textbf{Rate (\%)} & \textbf{IoU} & \textbf{dice} & \textbf{precision} & \textbf{recall} & \textbf{accuracy} & \textbf{panoptic quality} \\
\hline
1   & $0.669 \pm 0.208$ & $0.780 \pm 0.182$ & $0.891 \pm 0.113$ & $0.736 \pm 0.225$ & $0.818 \pm 0.151$ & $0.559 \pm 0.239$ \\
2   & $0.711 \pm 0.219$ & $0.807 \pm 0.196$ & $0.875 \pm 0.153$ & $0.788 \pm 0.228$ & $0.840 \pm 0.139$ & $0.615 \pm 0.248$ \\
3   & $0.725 \pm 0.207$ & $0.820 \pm 0.179$ & $0.888 \pm 0.124$ & $0.797 \pm 0.211$ & $0.846 \pm 0.144$ & $0.630 \pm 0.242$ \\
5   & $0.719 \pm 0.225$ & $0.811 \pm 0.203$ & $0.880 \pm 0.146$ & $0.791 \pm 0.235$ & $0.843 \pm 0.149$ & $0.627 \pm 0.255$ \\
10  & $0.728 \pm 0.206$ & $0.823 \pm 0.171$ & $0.903 \pm 0.101$ & $0.791 \pm 0.217$ & $0.851 \pm 0.140$ & $0.633 \pm 0.247$ \\
20  & $0.742 \pm 0.198$ & $0.834 \pm 0.164$ & $0.892 \pm 0.121$ & $0.810 \pm 0.207$ & $0.863 \pm 0.124$ & $0.651 \pm 0.238$ \\
30  & $0.743 \pm 0.203$ & $0.834 \pm 0.168$ & $0.894 \pm 0.123$ & $0.808 \pm 0.210$ & $0.864 \pm 0.124$ & $0.652 \pm 0.242$ \\
40  & $0.772 \pm 0.175$ & $0.858 \pm 0.144$ & $0.903 \pm 0.095$ & $0.837 \pm 0.182$ & $0.881 \pm 0.102$ & $0.687 \pm 0.210$ \\
50  & $0.763 \pm 0.189$ & $0.849 \pm 0.162$ & $0.914 \pm 0.079$ & $0.821 \pm 0.201$ & $0.881 \pm 0.099$ & $0.678 \pm 0.221$ \\
80  & $0.759 \pm 0.180$ & $0.849 \pm 0.143$ & $0.904 \pm 0.098$ & $0.822 \pm 0.186$ & $0.873 \pm 0.106$ & $0.669 \pm 0.219$ \\
100 & $0.771 \pm 0.182$ & $0.855 \pm 0.157$ & $0.909 \pm 0.079$ & $0.836 \pm 0.191$ & $0.885 \pm 0.100$ & $0.687 \pm 0.212$ \\
\hline
\end{tabular}
\label{tab:dq_full_metrics}
\end{center}
\end{table*}

\begin{figure}[t]
\centerline{\includegraphics[width=\linewidth]{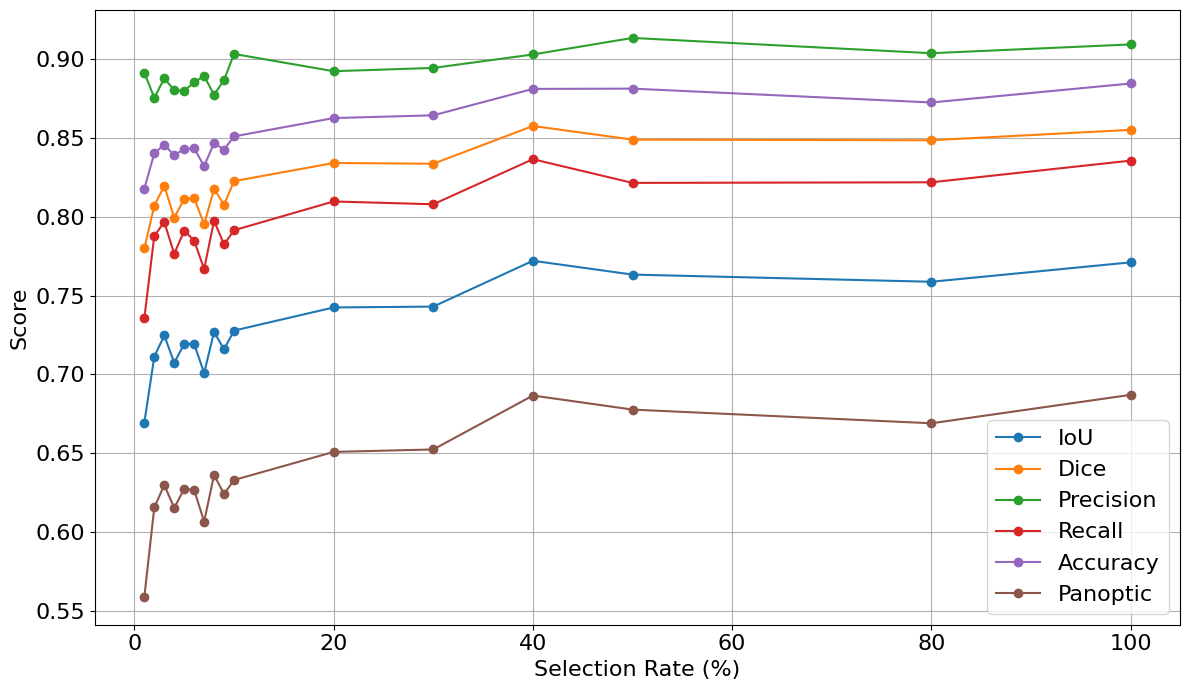}}
\caption{Image segmentation performance metrics under DQ across increasing training data percentages (1\%–100\%) in Cyto.}
\label{fig:redundancy}
\end{figure}

\subsection{Redundancy in Training Data}

Figure~\ref{fig:redundancy} and Table~\ref{tab:dq_full_metrics} show saturation in segmentation performance as the training subset grows, indicating substantial redundancy in the Cyto train set. The first $10\%$ already captures most essential information: precision increases only from $0.891\pm0.113$ (1\%) to $0.903\pm0.101$ (10\%), while accuracy plateaus around $0.851\pm0.140$. From 10--40\%, additional data yields modest gains on structure-sensitive metrics (dice: $0.823\pm0.171 \rightarrow 0.858\pm0.144$) via boundary refinement. Beyond 40\%, improvements are marginal (dice $\approx 0.858$ at 40\% vs.\ $0.855$ at 100\%; PQ varies narrowly), implying that $>60\%$ of training data is functionally redundant. In sum, 0--10\% provides core knowledge, 10--40\% auxiliary gains, and $>40\%$ little added value, supporting dataset quantization and coreset selection.

\begin{figure}[htbp]
\centerline{\includegraphics[width=\linewidth]{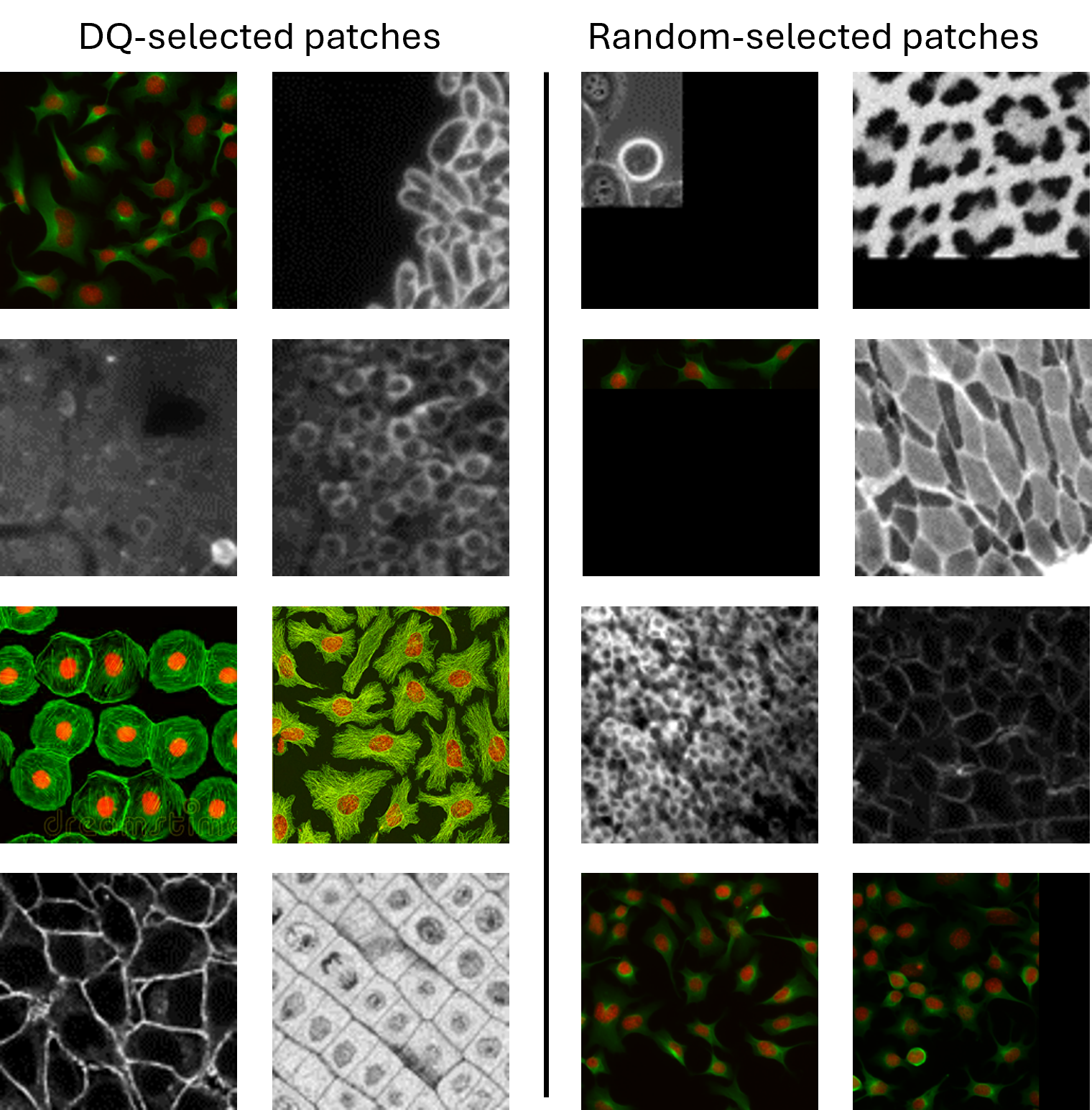}}
\caption{Example image patches selected by dataset quantization (left) and random sampling (right) at 1\% selection rate in Cyto trainset. DQ patches exhibit greater diversity in morphology and contrast.}
\label{fig:patch_examples}
\end{figure}

\begin{figure}[htbp]
\centerline{\includegraphics[width=\linewidth]{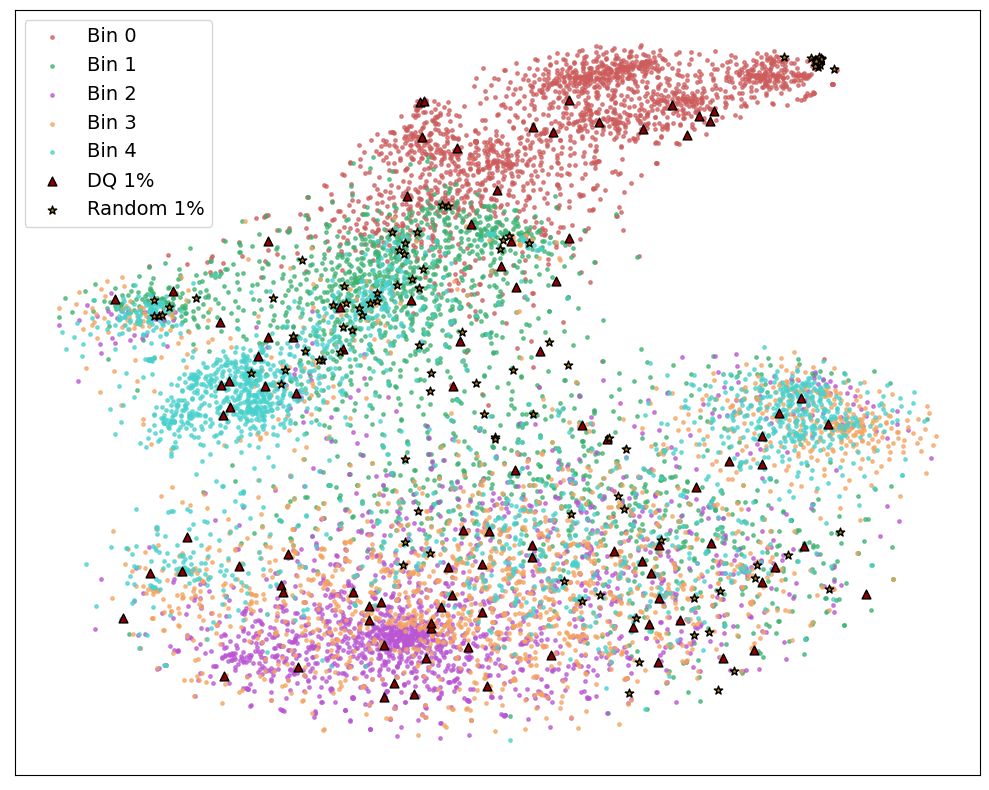}}
\caption{t-SNE projection of MAE features from Cyto trainset, grouped into 5 bins. DQ samples (red) show broader coverage in the latent feature space than Random (golden), indicating higher visual diversity.}
\label{fig:tsne_comparison}
\end{figure}

\begin{table*}[htbp]
\caption{DQ vs Random Comparison on \textbf{Cyto} (Train \& Test). Values are Mean ± Std over five runs.}
\begin{center}
\begin{tabular}{|c|c|c|c|c|c|c|}
\hline
\textbf{Rate (\%)} & \textbf{IoU (DQ)} & \textbf{IoU (Rand)} & \textbf{dice (DQ)} & \textbf{dice (Rand)} & \textbf{panoptic quality (DQ)} & \textbf{panoptic quality (Rand)} \\
\hline
1   & $0.669 \pm 0.208$ & $0.722 \pm 0.015$ & $0.780 \pm 0.182$ & $0.817 \pm 0.017$ & $0.559 \pm 0.239$ & $0.558 \pm 0.024$ \\
5   & $0.719 \pm 0.225$ & $0.728 \pm 0.007$ & $0.811 \pm 0.203$ & $0.820 \pm 0.007$ & $0.627 \pm 0.255$ & $0.627 \pm 0.006$ \\
10  & $0.728 \pm 0.206$ & $0.750 \pm 0.003$ & $0.823 \pm 0.171$ & $0.839 \pm 0.004$ & $0.633 \pm 0.247$ & $0.662 \pm 0.003$ \\
30  & $0.743 \pm 0.203$ & $0.771 \pm 0.006$ & $0.834 \pm 0.168$ & $0.856 \pm 0.005$ & $0.652 \pm 0.242$ & $0.687 \pm 0.005$ \\
50  & $0.763 \pm 0.189$ & $0.779 \pm 0.006$ & $0.849 \pm 0.162$ & $0.861 \pm 0.003$ & $0.678 \pm 0.221$ & $0.696 \pm 0.008$ \\
100 & $0.771 \pm 0.182$ & $0.771 \pm 0.182$ & $0.855 \pm 0.157$ & $0.855 \pm 0.157$ & $0.687 \pm 0.212$ & $0.687 \pm 0.212$ \\
\hline
\end{tabular}
\label{tab:dq_random_Cyto}
\end{center}
\end{table*}

\begin{table*}[htbp]
\caption{DQ vs Random Comparison on \textbf{MultiInst} (Train \& Test). Values are Mean ± Std over five runs.}
\begin{center}
\begin{tabular}{|c|c|c|c|c|c|c|}
\hline
\textbf{Rate (\%)} & \textbf{IoU (DQ)} & \textbf{IoU (Rand)} & \textbf{dice (DQ)} & \textbf{dice (Rand)} & \textbf{panoptic quality (DQ)} & \textbf{panoptic quality (Rand)} \\
\hline
1   & $0.338 \pm 0.331$ & $0.344 \pm 0.031$ & $0.427 \pm 0.325$ & $0.442 \pm 0.036$ & $0.248 \pm 0.346$ & $0.246 \pm 0.029$ \\
5   & $0.333 \pm 0.365$ & $0.346 \pm 0.025$ & $0.401 \pm 0.372$ & $0.435 \pm 0.031$ & $0.265 \pm 0.368$ & $0.257 \pm 0.021$ \\
10  & $0.348 \pm 0.357$ & $0.368 \pm 0.012$ & $0.424 \pm 0.361$ & $0.458 \pm 0.015$ & $0.272 \pm 0.363$ & $0.279 \pm 0.010$ \\
30  & $0.418 \pm 0.316$ & $0.373 \pm 0.022$ & $0.525 \pm 0.301$ & $0.461 \pm 0.032$ & $0.311 \pm 0.342$ & $0.285 \pm 0.013$ \\
50  & $0.407 \pm 0.342$ & $0.381 \pm 0.018$ & $0.498 \pm 0.341$ & $0.472 \pm 0.028$ & $0.315 \pm 0.354$ & $0.290 \pm 0.009$ \\
100 & $0.512 \pm 0.275$ & $0.512 \pm 0.275$ & $0.635 \pm 0.241$ & $0.635 \pm 0.241$ & $0.390 \pm 0.316$ & $0.390 \pm 0.316$ \\
\hline
\end{tabular}
\label{tab:dq_random_MultiInst}
\end{center}
\end{table*}

\subsection{Dataset Quantization for Coreset Selection}
To evaluate the effectiveness of dataset quantization (DQ) in selecting compact yet representative training subsets, we conduct a multi-level comparison with random sampling. This includes visual inspection of selected patches, structural analysis in latent space, and quantitative performance evaluation.

\textbf{Visual Analysis of Selected Patches}: Figure~\ref{fig:patch_examples} illustrates representative training patches from the Cyto dataset selected via DQ versus random sampling at the same subset ratio 1\%. Visually, DQ selected patches cover a wider variety of cell morphologies, staining patterns, and spatial configurations. In contrast, random sampling often includes visually repetitive or low information regions due to its stochastic nature. This suggests that DQ promotes structural and semantic diversity during subset construction.

\textbf{Structural Analysis in Latent Space}: To investigate feature level diversity, we extract latent representations of all patches using a pretrained Masked Autoencoder (MAE) and project the embeddings into 2D using t-SNE. As shown in Figure~\ref{fig:tsne_comparison}, DQ selected patches span a broader and more uniformly distributed region of the latent space. In contrast, random samples cluster in more localized zones, indicating potential redundancy. These results demonstrate that DQ provides more comprehensive coverage of the dataset’s intrinsic feature space.

\textbf{Quantitative Comparison with Random Sampling}: We further compare DQ and random selection quantitatively on the Cyto and MultiInst datasets at matched training rates. Performance metrics are summarized in Tables~\ref{tab:dq_random_Cyto} and~\ref{tab:dq_random_MultiInst}. While both methods yield comparable results, DQ exhibits slightly more stable performance, especially in low data regimes. On highly heterogeneous datasets like MultiInst, performance differences diminish due to inherent data diversity.

In summary, while DQ and random sampling achieve similar performance in image segmentation metrics, DQ provides advantages in feature diversity, determinism, and reproducibility. We position DQ as a simple yet interpretable coreset baseline for probing data efficiency, with potential for further enhancement via adaptive or learning based selection methods.

\subsection{Zero-Shot Generalization to MultiInst}

\begin{table}[htbp]
\caption{Zero-Shot Image Segmentation Performance on MultiInst Testset (Trained on Cyto, Mean ± Std).}
\centering
\begin{tabular}{|c|c|c|c|}
\hline
\textbf{Metric} & \textbf{Full Data (100\%)} & \textbf{DQ Subset (30\%)} \\
\hline
IoU & $0.419 \pm 0.291$ & $0.428 \pm 0.337$ \\
dice & $0.530 \pm 0.308$ & $0.516 \pm 0.365$ \\
precision & $0.810 \pm 0.213$ & $0.743 \pm 0.282$ \\
recall & $0.440 \pm 0.303$ & $0.453 \pm 0.354$ \\
accuracy & $0.820 \pm 0.128$ & $0.832 \pm 0.121$ \\
panoptic quality & $0.308 \pm 0.287$ & $0.341 \pm 0.319$ \\
\hline
\end{tabular}
\label{tab:zero_shot_MultiInst}
\end{table}

We assess cross domain generalization by applying Cellpose models that were trained solely on the Cyto dataset to the MultiInst testset without any finetuning. We compare two models: one was trained on the 100\% full dataset and one was trained on a 30\% DQ coreset.

Both models achieve nontrivial zero-shot performance, confirming moderate cross domain transferability. The full data model yields higher precision ($0.810 \pm 0.213$), while the DQ-trained model shows slightly better recall and accuracy, indicating more aggressive but less precise segmentation. This reflects typical domain shift behavior: MultiInst includes modalities and structures absent from Cyto, amplifying generalization challenges. Notably, the DQ model exhibits greater performance variance (higher standard deviations in IoU and dice), suggesting that aggressive data reduction can increase sensitivity under distributional shifts.

DQ supports efficient training and retains reasonable zero-shot generalization capability, but its increased variance under domain shifts cautions against its use in critical settings where consistency is required. Future work may explore combining DQ with uncertainty or diversity-based sampling to enhance both robustness and transferability. This reflects a balance between efficiency and robustness, particularly under unseen domain shift.

\subsection{Forgetting During Transfer}

\begin{table*}[t]
\caption{Cross Domain finetuning Results on Cyto and Histo (Mean ± Std)}
\begin{center}
\renewcommand{\arraystretch}{1.2}
\setlength{\tabcolsep}{4pt}
\begin{tabular}{|l|l|c|c|c|c|c|c|}
\hline
\textbf{Training Strategy} & \textbf{Test Domain} & \textbf{IoU} & \textbf{dice} & \textbf{precision} & \textbf{recall} & \textbf{accuracy} & \textbf{panoptic quality} \\
\hline
Cyto only 
& Cyto & $0.771 \pm 0.182$ & $0.855 \pm 0.157$ & $0.909 \pm 0.079$ & $0.836 \pm 0.191$ & $0.885 \pm 0.100$ & $0.687 \pm 0.212$ \\
& Histo  & $0.358 \pm 0.134$ & $0.512 \pm 0.169$ & $0.520 \pm 0.170$ & $0.530 \pm 0.221$ & $0.804 \pm 0.044$ & $0.204 \pm 0.104$ \\
\hline
Histo only 
& Cyto & $0.081 \pm 0.128$ & $0.129 \pm 0.185$ & $0.486 \pm 0.388$ & $0.086 \pm 0.134$ & $0.499 \pm 0.257$ & $0.034 \pm 0.072$ \\
& Histo  & $0.688 \pm 0.077$ & $0.812 \pm 0.060$ & $0.792 \pm 0.093$ & $0.843 \pm 0.055$ & $0.919 \pm 0.019$ & $0.563 \pm 0.094$ \\
\hline
Cyto + Histo 
& Cyto & $0.763 \pm 0.186$ & $0.850 \pm 0.153$ & $0.898 \pm 0.098$ & $0.832 \pm 0.192$ & $0.871 \pm 0.125$ & $0.676 \pm 0.223$ \\
& Histo  & $0.688 \pm 0.046$ & $0.814 \pm 0.032$ & $0.805 \pm 0.048$ & $0.829 \pm 0.068$ & $0.920 \pm 0.021$ & $0.562 \pm 0.059$ \\
\hline
Pretrain: Cyto, Finetune: Histo 
& Cyto & $0.047 \pm 0.083$ & $0.080 \pm 0.129$ & $0.417 \pm 0.404$ & $0.050 \pm 0.088$ & $0.474 \pm 0.270$ & $0.014 \pm 0.038$ \\
& Histo  & $0.692 \pm 0.053$ & $0.817 \pm 0.037$ & $0.804 \pm 0.056$ & $0.837 \pm 0.069$ & $0.922 \pm 0.018$ & $0.567 \pm 0.068$ \\
\hline
Pretrain: Histo, Finetune: Cyto 
& Cyto & $0.771 \pm 0.180$ & $0.856 \pm 0.149$ & $0.900 \pm 0.089$ & $0.841 \pm 0.185$ & $0.882 \pm 0.101$ & $0.686 \pm 0.216$ \\
& Histo  & $0.402 \pm 0.124$ & $0.561 \pm 0.151$ & $0.525 \pm 0.129$ & $0.628 \pm 0.212$ & $0.807 \pm 0.043$ & $0.243 \pm 0.102$ \\
\hline
\end{tabular}
\label{tab:transfer_full}
\end{center}
\end{table*}

\begin{figure}[htbp]
\centerline{\includegraphics[width=\linewidth]{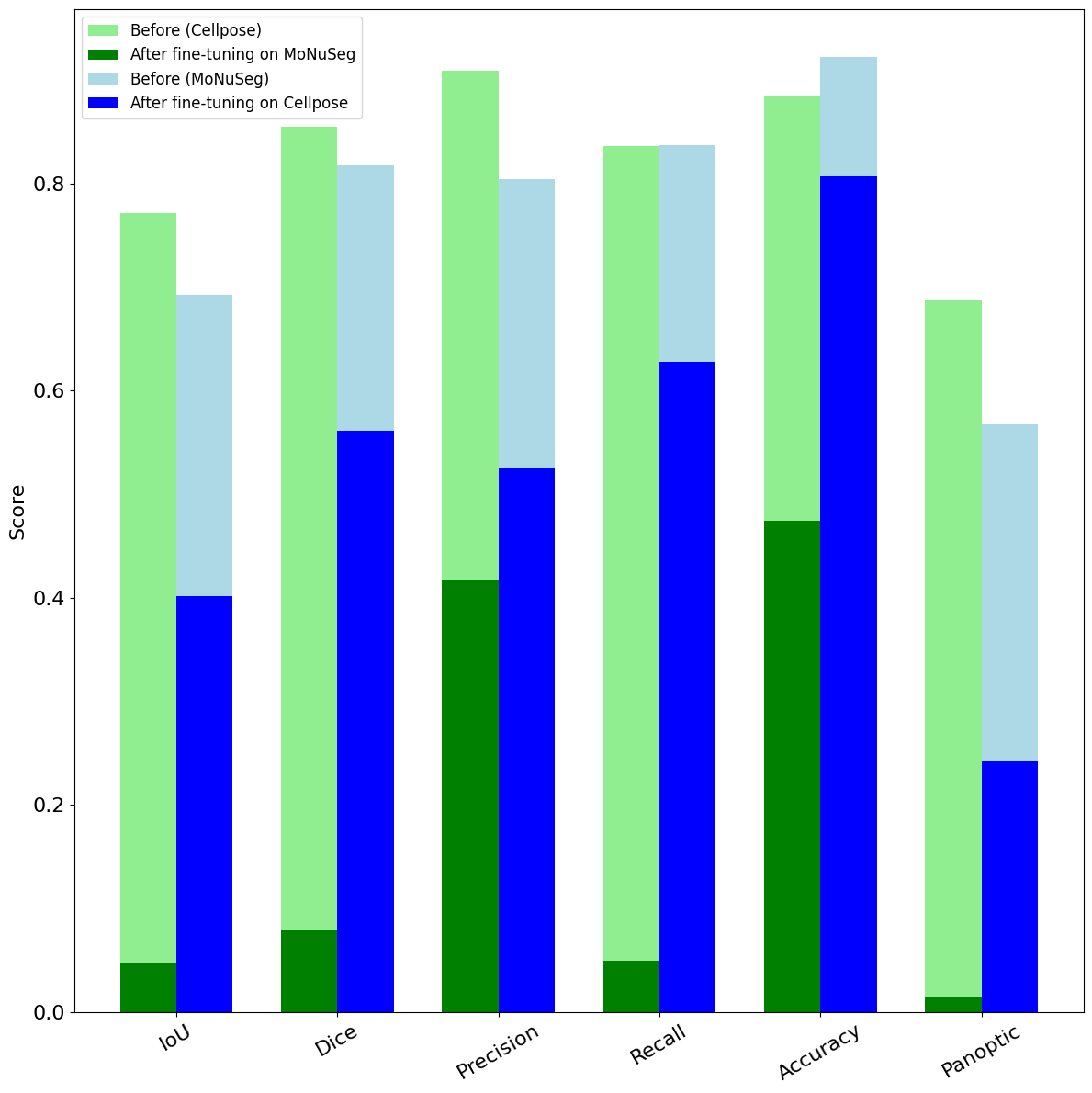}}
\caption{Performance degradation before and after cross-domain finetuning across six evaluation metrics. Catastrophic forgetting is observed in Cyto-to-Histo transfer, with source domain metrics dropping sharply. The reverse transfer results in milder forgetting.}
\label{fig:forgetting}
\end{figure}

We evaluate bidirectional transfer between Cyto and Histo using multiple training strategies (Table~\ref{tab:transfer_full}, Fig.~\ref{fig:forgetting}). finetuning Cyto pretrained models on Histo (\textit{Cyto $\rightarrow$ Histo}) leads to severe forgetting: IoU drops from ($0.771 \pm 0.182$) to ($0.047 \pm 0.083$), dice from ($0.855 \pm 0.157$) to ($0.080 \pm 0.129$), and panoptic quality from ($0.687 \pm 0.212$) to ($0.014 \pm 0.038$), indicating near total loss of source task knowledge.

In contrast, reverse transfer (\textit{Histo $\rightarrow$ Cyto}) results in milder forgetting, with Cyto performance largely retained, shows dice ($0.561 \pm 0.151$), panoptic quality ($0.243 \pm 0.102$). This asymmetry suggests that Cyto’s generalist features are more transferable, while Histo’s narrow domain features are more fragile.

Joint training (\textit{Cyto+Histo}) mitigates forgetting, achieving balanced performance across both domains. These results highlight the limitations of naive finetuning and point to the need for continual learning approaches, such as replay, regularization, or modular architectures, to ensure knowledge retention under domain shift.

\subsection{Transfer Robustness of DQ Trained Models}

We examine whether training on quantized subsets improves robustness to forgetting. Models are first trained on varying fractions of Cyto trainset (1\%--50\%, via DQ), then finetuned on full Histo trainset. We evaluate forgetting by measuring transfer performance on the original Cyto testset.

As shown in Table~\ref{tab:forgetting_dq}, all DQ trained models suffer substantial degradation after transfer, with IoU dropping from $0.763$ (100\% baseline) to as low as $0.042$--$0.086$, and dice dropping by over 0.7 points in some cases. Surprisingly, larger DQ subsets (e.g., 50\%) do not confer clear resilience compared to smaller ones.

These results suggest that dataset compression alone, while beneficial for efficient training, does not mitigate forgetting. Instead, the retention gap calls for more principled strategies, such as replay or continual learning mechanisms, to preserve source task knowledge during domain adaptation.

\begin{table*}[htbp]
\caption{Image Segmentation Performance of DQ trained models on Cyto testset after finetuning on Histo (Mean ± Std).}
\begin{center}
\begin{tabular}{|c|c|c|c|c|c|c|}
\hline
\textbf{DQ Rate (\%)} & \textbf{IoU} & \textbf{dice} & \textbf{precision} & \textbf{recall} & \textbf{accuracy} & \textbf{panoptic quality} \\
\hline
1   & 0.060 ± 0.094 & 0.100 ± 0.144 & 0.446 ± 0.389 & 0.064 ± 0.101 & 0.480 ± 0.264 & 0.019 ± 0.047 \\
5   & 0.086 ± 0.115 & 0.142 ± 0.168 & 0.500 ± 0.383 & 0.095 ± 0.127 & 0.490 ± 0.255 & 0.031 ± 0.066 \\
10  & 0.042 ± 0.070 & 0.074 ± 0.112 & 0.447 ± 0.374 & 0.045 ± 0.075 & 0.476 ± 0.272 & 0.011 ± 0.030 \\
30  & 0.063 ± 0.100 & 0.105 ± 0.152 & 0.434 ± 0.367 & 0.067 ± 0.105 & 0.484 ± 0.258 & 0.022 ± 0.050 \\
50  & 0.042 ± 0.077 & 0.072 ± 0.119 & 0.466 ± 0.399 & 0.044 ± 0.080 & 0.476 ± 0.271 & 0.012 ± 0.036 \\
\hline
\end{tabular}
\label{tab:forgetting_dq}
\end{center}
\end{table*}

\subsection{Multi-Stage Transfer Order Matters}

\begin{table*}[t]
\caption{Image Segmentation Performance of Multi-Stage Transfer Paths Across Three Test Domains.}
\label{tab:multi_stage_transfer}
\centering
\resizebox{\textwidth}{!}{%
\begin{tabular}{|l|c|c|c|c|c|c|c|c|c|c|c|c|c|c|c|c|c|c|}
\hline
\multirow{2}{*}{Training Path} & \multicolumn{6}{c|}{\textbf{Test: Cyto}} & \multicolumn{6}{c|}{\textbf{Test: Histo}} & \multicolumn{6}{c|}{\textbf{Test: MultiInst}} \\
\cline{2-19}
& IoU & dice & prec. & recall & acc. & PQ & IoU & dice & prec. & recall & acc. & PQ & IoU & dice & prec. & recall & acc. & PQ \\
\hline
Cyto only & 0.768 & 0.853 & 0.903 & 0.833 & 0.879 & 0.684 & 0.358 & 0.512 & 0.520 & 0.530 & 0.804 & 0.204 & 0.355 & 0.439 & 0.786 & 0.367 & 0.811 & 0.272 \\
Cyto$\rightarrow$Histo & 0.047 & 0.080 & 0.417 & 0.050 & 0.474 & 0.014 & \textbf{0.692} & \textbf{0.817} & \textbf{0.804} & \textbf{0.837} & \textbf{0.922} & \textbf{0.567} & 0.257 & 0.350 & 0.850 & 0.262 & 0.725 & 0.164 \\
Path A: Cyto$\rightarrow$Histo$\rightarrow$MultiInst & 0.465 & 0.594 & 0.848 & 0.523 & 0.669 & 0.337 & 0.232 & 0.371 & 0.343 & 0.429 & 0.703 & 0.093 & \textbf{0.479} & \textbf{0.594} & \textbf{0.928} & \textbf{0.489} & \textbf{0.823} & \textbf{0.365} \\
Cyto$\rightarrow$MultiInst & 0.376 & 0.492 & 0.782 & 0.420 & 0.615 & 0.260 & 0.226 & 0.363 & 0.343 & 0.410 & 0.708 & 0.089 & 0.523 & 0.638 & 0.908 & 0.537 & 0.829 & 0.408 \\
Path B: Cyto$\rightarrow$MultiInst$\rightarrow$Histo & 0.142 & 0.214 & 0.545 & 0.162 & 0.505 & 0.069 & \textbf{0.682} & \textbf{0.810} & \textbf{0.765} & \textbf{0.865} & \textbf{0.915} & \textbf{0.555} & 0.244 & 0.341 & 0.797 & 0.251 & 0.741 & 0.146 \\
MultiInst only & 0.420 & 0.546 & 0.837 & 0.468 & 0.646 & 0.293 & 0.218 & 0.351 & 0.332 & 0.396 & 0.707 & 0.085 & 0.512 & 0.635 & 0.920 & 0.525 & 0.826 & 0.390 \\
MultiInst$\rightarrow$Cyto & \textbf{0.762} & \textbf{0.848} & \textbf{0.915} & \textbf{0.820} & \textbf{0.876} & \textbf{0.677} & 0.357 & 0.512 & 0.504 & 0.550 & 0.795 & 0.202 & 0.398 & 0.490 & 0.804 & 0.416 & 0.822 & 0.306 \\
Path C: MultiInst$\rightarrow$Cyto$\rightarrow$Histo & 0.160 & 0.243 & 0.578 & 0.179 & 0.511 & 0.077 & \textbf{0.680} & \textbf{0.807} & \textbf{0.764} & \textbf{0.864} & \textbf{0.914} & \textbf{0.552} & 0.295 & 0.411 & 0.862 & 0.309 & 0.749 & 0.178 \\
\hline
\end{tabular}
}
\end{table*}

As discussed in Section~\ref{sec:methodology}, we investigate how sequential domain adaptation order affects model retention and final performance. Three training paths are compared across Cyto, Histo, and MultiInst datasets.

Table~\ref{tab:multi_stage_transfer} shows that training order significantly impacts both forgetting and generalization. Path~A (Cyto~$\rightarrow$~Histo~$\rightarrow$~MultiInst) yields the best MultiInst performance but forgets prior tasks. Path~B ends with Histo and excels there but loses Cyto knowledge. In contrast, Path~C (MultiInst~$\rightarrow$~Cyto~$\rightarrow$~Histo) achieves more balanced performance across all domains, indicating better retention.

We can get several observations: (1) Early exposure to simple domains like Cyto often leads to their knowledge being forgotten. (2) Starting with diverse domains like MultiInst helps build transferable representations. (3) Inserting generalist domains midstage improves retention without harming downstream performance.

These findings highlight the importance of training curricula in multidomain biomedical image segmentation and suggest that careful sequencing can mitigate forgetting without additional architectural changes.

\textbf{Key findings:}

\begin{itemize}
    \item \textbf{Path A} achieves the best performance on the final task (MultiInst) but suffers from moderate forgetting on Cyto and Histo.
    \item \textbf{Path B} yields the best Histo performance, suggesting that placing Histo last in the sequence helps preserve its target specific features. However, Cyto knowledge is significantly forgotten.
    \item \textbf{Path C} strikes a balance: it recovers strong Cyto performance via intermediate finetuning while still achieving high MultiInst and Histo scores, indicating better retention and generalization.
\end{itemize}

These results highlight that the order of domain exposure impacts both model forgetting and final generalization. We observe that:
\begin{itemize}
    \item Simpler and more homogeneous domains like Cyto are more prone to being forgotten when exposed early in the training sequence.
    \item Datasets with high variability and stronger semantic signals like MultiInst serve better as base representations.
    \item Transfer robustness depends not only on model capacity but also on the compatibility and visual diversity between source and target domains.
\end{itemize}

This experiment motivates future work on curriculum aware and rehearsal based continual image segmentation strategies, and highlights the importance of training order when working with heterogeneous biomedical datasets.

\subsection{Simple Replay Strategy for Forgetting Mitigation}

\begin{table*}[htbp]
\caption{Replay experiments: combining Cyto DQ subsets (1--100\%) with full Histo training. Transfer performance reported on both test sets (Mean ± Std).}
\centering
\resizebox{\textwidth}{!}{
\begin{tabular}{|l|cccccc|cccccc|}
\hline
\multirow{2}{*}{\textbf{Finetune Stages}} & \multicolumn{6}{c|}{\textbf{Cyto Testset}} & \multicolumn{6}{c|}{\textbf{Histo Testset}} \\
\cline{2-13}
& IoU & dice & precision & recall & accuracy & panoptic & IoU & dice & precision & recall & accuracy & panoptic \\
\hline
0\% Cyto + 100\% Histo & 0.047 ± 0.083 & 0.080 ± 0.129 & 0.417 ± 0.404 & 0.050 ± 0.088 & 0.474 ± 0.270 & 0.014 ± 0.038 & 0.692 ± 0.053 & 0.817 ± 0.037 & 0.804 ± 0.056 & 0.837 ± 0.069 & 0.922 ± 0.018 & 0.567 ± 0.068 \\
1\% Cyto + 100\% Histo & 0.590 ± 0.261 & 0.701 ± 0.259 & 0.869 ± 0.185 & 0.648 ± 0.285 & 0.774 ± 0.194 & 0.479 ± 0.272 & \textbf{0.696 ± 0.058} & \textbf{0.820 ± 0.043} & 0.807 ± 0.038 & \textbf{0.838 ± 0.081} & \textbf{0.923 ± 0.019} & \textbf{0.573 ± 0.074} \\
5\% Cyto + 100\% Histo & 0.720 ± 0.219 & 0.813 ± 0.195 & 0.883 ± 0.130 & 0.791 ± 0.230 & 0.844 ± 0.156 & 0.626 ± 0.250 & 0.693 ± 0.055 & 0.818 ± 0.039 & 0.790 ± 0.051 & 0.852 ± 0.064 & 0.920 ± 0.020 & 0.569 ± 0.070 \\
10\% Cyto + 100\% Histo & 0.738 ± 0.198 & 0.831 ± 0.164 & 0.899 ± 0.110 & 0.803 ± 0.208 & 0.859 ± 0.133 & 0.644 ± 0.237 & 0.676 ± 0.071 & 0.804 ± 0.054 & 0.784 ± 0.054 & 0.833 ± 0.091 & 0.916 ± 0.021 & 0.547 ± 0.088 \\
30\% Cyto + 100\% Histo & 0.741 ± 0.215 & 0.829 ± 0.190 & 0.896 ± 0.121 & 0.805 ± 0.226 & 0.862 ± 0.135 & 0.654 ± 0.246 & 0.680 ± 0.063 & 0.808 ± 0.047 & 0.792 ± 0.048 & 0.832 ± 0.087 & 0.917 ± 0.021 & 0.552 ± 0.079 \\
50\% Cyto + 100\% Histo & 0.762 ± 0.181 & 0.850 ± 0.148 & 0.908 ± 0.098 & 0.821 ± 0.190 & 0.876 ± 0.108 & 0.674 ± 0.218 & 0.673 ± 0.068 & 0.803 ± 0.051 & \textbf{0.808 ± 0.042} & 0.804 ± 0.090 & 0.918 ± 0.021 & 0.544 ± 0.084 \\
100\% Cyto + 100\% Histo& \textbf{0.765 ± 0.186} & \textbf{0.851 ± 0.153} & \textbf{0.911 ± 0.086} & \textbf{0.823 ± 0.197} & \textbf{0.881 ± 0.105} & \textbf{0.678 ± 0.222} & 0.310 ± 0.123 & 0.459 ± 0.155 & 0.463 ± 0.129 & 0.488 ± 0.231 & 0.781 ± 0.043 & 0.160 ± 0.095 \\
\hline
\end{tabular}
}
\label{tab:replay_results}
\end{table*}

We assess whether simple replay can alleviate forgetting by combining a partial Cyto trainset (1\%–100\%, via DQ) with full Histo trainset finetuning. As shown in Table~\ref{tab:replay_results}, replaying just 5--10\% of Cyto substantially recovers source performance like dice improves from $0.08 \pm 0.13$ to $0.83 \pm 0.16$, while Histo accuracy remains stable ($\sim$0.80).

However, using 100\% Cyto unexpectedly degrades Histo results (dice drops to $0.46$), suggesting that excessive replay may harm target domain learning.

These findings highlight that even simple DQ-based replay can effectively mitigate forgetting. While more sophisticated strategies like regularization and dynamic memory may offer further gains, our goal is to provide a lightweight, reproducible baseline that motivates future work.

\section{Summary, Discussion and Outlook}

This study tackles two core challenges in biomedical image segmentation: dataset reduction and cross-domain transfer. Using Cellpose as a testbed, we analyze how much training data is needed, how domain shifts cause forgetting, and how strategies like dataset quantization (DQ), replay, and training order impact generalization and retention across diverse domains. By using Cellpose as a representative model, we derive the following key insights:

\begin{itemize}
    \item \textbf{Redundancy and Compact Subsets:} Strong image segmentation performance can be achieved with only 10--40\% of the Cyto training data. Dataset quantization (DQ) effectively identifies structurally diverse, representative patches, enabling efficient training with minimal performance loss.

    \item \textbf{DQ vs. Random Sampling:} While DQ and random subsets yield in metrics, DQ offers superior feature diversity, determinism, and reproducibility, making it a practical coreset selection strategy in data-centric pipelines.

    \item \textbf{Cross Domain Transferability:} DQ trained models generalize well to unseen domains, particularly in recall and panoptic quality, suggesting that compact yet diverse training subsets can support transferable representations.

    \item \textbf{Forgetting and Replay:} Naive finetuning causes significant forgetting, especially from generalist to specialist domains. Incorporating 5--10\% of the source data as replay during target finetuning substantially recovers source performance. Interestingly, replaying the full dataset may hurt target performance, highlighting the importance of selective replay.

    \item \textbf{Impact of Training Order:} In multi-stage setups, the order of domain exposure critically affects final performance. Starting with heterogeneous data (e.g., MultiInst), followed by generalist replay (Cyto), and ending with domain specific finetuning (Histo) yields the best balance between generalization and retention.
\end{itemize}

These findings reveal that image segmentation quality is influenced not only by model architecture, but also by the composition, diversity, and sequencing of training data. While DQ improves training efficiency, it does not inherently prevent forgetting. Simple strategies such as selective replay can serve as lightweight yet effective solutions.

We identify three key limitations in current pipelines: lack of retention mechanisms, monolithic model design, and uneven annotation quality. To address these, we recommend the following directions:

\begin{itemize}
    \item \textbf{Retention aware learning:} Combine DQ based replay with continual learning methods such as knowledge distillation or memory based regularization.
    \item \textbf{Model design:} Use modular architectures to separate domain-specific components and reduce interference.
    \item \textbf{Curriculum aware training:} Leverage domain sequencing to guide generalization and mitigate forgetting.
    \item \textbf{Advanced subset selection:} Enhance DQ with uncertainty or diversity metrics to refine coreset quality.
\end{itemize}

In conclusion, this work highlights the value of data-centric design in biomedical image segmentation. By coupling compact training subsets with simple retention strategies, we can build robust, efficient, and scalable image segmentation systems for real-world biomedical applications.

\section*{Acknowledgment}

This work was supported by Federal Ministry of Research, Technology and Space (Bundesministerium für Forschung, Technologie und Raumfahrt, BMFTR) under the funding reference 161L0272 and the “Ministerium für Kultur und Wissenschaft des Landes Nordrhein-Westfalen”. J. Chen was also partially supported by NFDI4Bioimage, funded by the German Research Foundation (DFG) within the framework of the NFDI-project number 501864659.


\bibliographystyle{IEEEtran}

\begin{thebibliography}{10}
\providecommand{\url}[1]{#1}
\csname url@samestyle\endcsname
\providecommand{\newblock}{\relax}
\providecommand{\bibinfo}[2]{#2}
\providecommand{\BIBentrySTDinterwordspacing}{\spaceskip=0pt\relax}
\providecommand{\BIBentryALTinterwordstretchfactor}{4}
\providecommand{\BIBentryALTinterwordspacing}{\spaceskip=\fontdimen2\font plus
\BIBentryALTinterwordstretchfactor\fontdimen3\font minus \fontdimen4\font\relax}
\providecommand{\BIBforeignlanguage}[2]{{%
\expandafter\ifx\csname l@#1\endcsname\relax
\typeout{** WARNING: IEEEtran.bst: No hyphenation pattern has been}%
\typeout{** loaded for the language `#1'. Using the pattern for}%
\typeout{** the default language instead.}%
\else
\language=\csname l@#1\endcsname
\fi
#2}}
\providecommand{\BIBdecl}{\relax}
\BIBdecl

\bibitem{stringer2021cellpose}
C.~Stringer, T.~Wang, M.~Michaelos, and M.~Pachitariu, ``Cellpose: a generalist algorithm for cellular segmentation,'' \emph{Nature methods}, vol.~18, no.~1, pp. 100--106, 2021.

\bibitem{cao2025rethinking}
J.~Cao, J.~Wenzel, S.~Zhang, J.~Lampe, H.~Wang, J.~Yao, Z.~Zhang, S.~Zhao, Y.~Zhou, C.~Chen, and M.~Schwaninger, ``Rethinking deep learning in bioimaging through a data centric lens,'' \emph{npj Imaging}, vol.~3, no.~1, p.~29, 2025.

\bibitem{NeurIPS-CellSeg}
J.~Ma, R.~Xie, S.~Ayyadhury, C.~Ge, A.~Gupta, R.~Gupta, S.~Gu, Y.~Zhang, G.~Lee, J.~Kim, W.~Lou, H.~Li, E.~Upschulte, T.~Dickscheid, J.~G. de~Almeida, Y.~Wang, L.~Han, X.~Yang, M.~Labagnara, V.~Gligorovski, M.~Scheder, S.~J. Rahi, C.~Kempster, A.~Pollitt, L.~Espinosa, T.~Mignot, J.~M. Middeke, J.-N. Eckardt, W.~Li, Z.~Li, X.~Cai, B.~Bai, N.~F. Greenwald, D.~Van~Valen, E.~Weisbart, B.~A. Cimini, T.~Cheung, O.~Brück, G.~D. Bader, and B.~Wang, ``The multi-modality cell segmentation challenge: Towards universal solutions,'' \emph{Nature Methods}, vol.~21, pp. 1103--1113, 2024.

\bibitem{pachitariu2022cellpose}
M.~Pachitariu and C.~Stringer, ``Cellpose 2.0: how to train your own model,'' \emph{Nature methods}, vol.~19, no.~12, pp. 1634--1641, 2022.

\bibitem{stringer2025cellpose3}
C.~Stringer and M.~Pachitariu, ``Cellpose3: one-click image restoration for improved cellular segmentation,'' \emph{Nature Methods}, pp. 1--8, 2025.

\bibitem{pachitariu2025cellpose}
M.~Pachitariu, M.~Rariden, and C.~Stringer, ``Cellpose-sam: superhuman generalization for cellular segmentation,'' \emph{bioRxiv}, pp. 2025--04, 2025.

\bibitem{zhao2025active}
S.~Zhao, Y.~Zhou, and J.~Chen, ``Active learning pipeline for biomedical image instance segmentation with minimal human intervention,'' in \emph{Bildverarbeitung für die Medizin (BVM) Workshop}.\hskip 1em plus 0.5em minus 0.4em\relax Wiesbaden: Springer Fachmedien Wiesbaden, 2025.

\bibitem{nath2020diminishing}
V.~Nath, D.~Yang, B.~A. Landman, D.~Xu, and H.~R. Roth, ``Diminishing uncertainty within the training pool: Active learning for medical image segmentation,'' \emph{IEEE Transactions on Medical Imaging}, vol.~40, no.~10, pp. 2534--2547, 2020.

\bibitem{gaillochet2023active}
M.~Gaillochet, C.~Desrosiers, and H.~Lombaert, ``Active learning for medical image segmentation with stochastic batches,'' \emph{Medical Image Analysis}, vol.~90, p. 102958, 2023.

\bibitem{wang2024comprehensive}
H.~Wang, Q.~Jin, S.~Li, S.~Liu, M.~Wang, and Z.~Song, ``A comprehensive survey on deep active learning in medical image analysis,'' \emph{Medical Image Analysis}, p. 103201, 2024.

\bibitem{yang2023towards}
Y.~Yang, H.~Kang, and B.~Mirzasoleiman, ``Towards sustainable learning: Coresets for data-efficient deep learning,'' in \emph{International Conference on Machine Learning}.\hskip 1em plus 0.5em minus 0.4em\relax PMLR, 2023, pp. 39\,314--39\,330.

\bibitem{hong2024evolution}
Y.~Hong, X.~Zhang, X.~Zhang, and J.~T. Zhou, ``Evolution-aware variance (eva) coreset selection for medical image classification,'' in \emph{Proceedings of the 32nd ACM International Conference on Multimedia}, 2024, pp. 301--310.

\bibitem{qazi2024continual}
M.~A. Qazi, A.~U.~R. Hashmi, S.~Sanjeev, I.~Almakky, N.~Saeed, C.~Gonzalez, and M.~Yaqub, ``Continual learning in medical imaging: A survey and practical analysis,'' \emph{arXiv preprint arXiv:2405.13482}, 2024.

\bibitem{yoon2023domain}
J.~S. Yoon, K.~Oh, Y.~Shin, M.~A. Mazurowski, and H.-I. Suk, ``Domain generalization for medical image analysis: A survey,'' \emph{arXiv preprint arXiv:2310.08598}, 2023.

\bibitem{garg2022multi}
P.~Garg, R.~Saluja, V.~N. Balasubramanian, C.~Arora, A.~Subramanian, and C.~Jawahar, ``Multi-domain incremental learning for semantic segmentation,'' in \emph{Proceedings of the IEEE/CVF Winter Conference on Applications of Computer Vision}, 2022, pp. 761--771.

\bibitem{qu2025recent}
H.~Qu, H.~Rahmani, L.~Xu, B.~Williams, and J.~Liu, ``Recent advances of continual learning in computer vision: An overview,'' \emph{IET Computer Vision}, vol.~19, no.~1, p. e70013, 2025.

\bibitem{kumar2019multi}
N.~Kumar, R.~Verma, D.~Anand, Y.~Zhou, O.~F. Onder, E.~Tsougenis, H.~Chen, P.-A. Heng, J.~Li, Z.~Hu \emph{et~al.}, ``A multi-organ nucleus segmentation challenge,'' \emph{IEEE transactions on medical imaging}, vol.~39, no.~5, pp. 1380--1391, 2019.

\bibitem{zhou2023dataset}
D.~Zhou, K.~Wang, J.~Gu, X.~Peng, D.~Lian, Y.~Zhang, Y.~You, and J.~Feng, ``Dataset quantization,'' in \emph{Proceedings of the IEEE/CVF International Conference on Computer Vision}, 2023, pp. 17\,205--17\,216.

\bibitem{he2022masked}
K.~He, X.~Chen, S.~Xie, Y.~Li, P.~Doll{\'a}r, and R.~Girshick, ``Masked autoencoders are scalable vision learners,'' in \emph{Proceedings of the IEEE/CVF conference on computer vision and pattern recognition}, 2022, pp. 16\,000--16\,009.

\bibitem{rolnick2019experience}
D.~Rolnick, A.~Ahuja, J.~Schwarz, T.~Lillicrap, and G.~Wayne, ``Experience replay for continual learning,'' in \emph{Advances in Neural Information Processing Systems}, vol.~32.\hskip 1em plus 0.5em minus 0.4em\relax Curran Associates, Inc., 2019.

\end{thebibliography}

\end{document}